\documentclass{article} 

\usepackage{iclr2027_conference,times}

\usepackage{amsmath,amsfonts,bm}

\def\eqref#1{equation~\ref{#1}}

\def\1{\bm{1}}

\DeclareMathAlphabet{\mathsfit}{\encodingdefault}{\sfdefault}{m}{sl}
\SetMathAlphabet{\mathsfit}{bold}{\encodingdefault}{\sfdefault}{bx}{n}

\usepackage{amsmath}
\usepackage{amssymb}
\usepackage{bm}

\usepackage[utf8]{inputenc} 
\usepackage[T1]{fontenc}    
\usepackage[hypertexnames=false,hyperfootnotes=false]{hyperref}
\usepackage{url}            
\usepackage{booktabs}       
\usepackage{amsfonts}       
\usepackage{float}          
\usepackage{nicefrac}       
\usepackage{microtype}      
\usepackage[dvipsnames]{xcolor}
\usepackage[capitalize,noabbrev]{cleveref}
\usepackage{enumitem}
\usepackage{placeins}

\usepackage{lineno}

\definecolor{cornflowerblue}{rgb}{0.39, 0.58, 0.93}
\hypersetup{
  colorlinks=true,
  linkcolor=cornflowerblue,
  filecolor=magenta,
  urlcolor=teal,
  citecolor=cornflowerblue,
}

\usepackage{tcolorbox}

\tcbuselibrary{skins}

\newtcolorbox{summarybox}[1][]{
  enhanced,
  colback=cornflowerblue!6!white,
  colframe=cornflowerblue!70!black,
  colbacktitle=cornflowerblue!20!white,
  coltitle=cornflowerblue!40!black,
  fonttitle=\bfseries,
  title=#1,
  boxrule=0.4pt,
  titlerule=0pt,                
  arc=2pt,
  left=8pt, right=8pt,
  top=4pt, bottom=6pt,
  toptitle=3pt, bottomtitle=3pt,  
  boxsep=0pt,
  before skip=10pt, after skip=10pt,
}

\newcommand{\blfootnote}[1]{%
  \begingroup
  \renewcommand{\thefootnote}{}%
  \footnotetext{#1}%
  \endgroup
}

\title{The Role of Quantization in Loss Spikes}

\author{%
  \begin{tabular}{l}
    Zaiqian Chen\textsuperscript{1}\thanks{Corresponding author: \texttt{zc2666@columbia.edu}.}
    \quad Aditya Somasundaram\textsuperscript{2}
    \quad Sophie Li\textsuperscript{1}
    \quad Sean McLeish\textsuperscript{3}
    \\
    John Kirchenbauer\textsuperscript{4}
    \quad Tom Goldstein\textsuperscript{3}
    \quad Micah Goldblum\textsuperscript{1}
  \end{tabular}%
}

\iclrfinalcopy
\begin{document}

\maketitle
\blfootnote{%
  \parbox[t]{\dimexpr\linewidth-1.8em\relax}{%
    \textsuperscript{1}\,Columbia University \quad
    \textsuperscript{2}\,Mohamed bin Zayed University of Artificial Intelligence\\
    \textsuperscript{3}\,University of Maryland \quad
    \textsuperscript{4}\,Vector Institute.%
  }%
}
\lhead{Preprint}

\begin{abstract}

  Loss spikes in language model pretraining can waste substantial compute and derail otherwise healthy training runs. A primary driver of these instabilities is the growth and divergence of output logits in the final language modeling (LM) head. In this work, we study a particularly costly form of this instability associated with output logit growth: delayed, terminal divergences that emerge only after prolonged training, which we call time-bomb spikes. We find that these instabilities are preceded by rapidly growing quantization error in the LM head, which eventually culminates in a loss spike. Crucially, reducing numerical precision accelerates time-bomb spikes, while increasing precision improves stability. We then propose a simple yet effective mitigation that replaces AdamW on the LM head with SGD with momentum, suppressing logit growth and improving stability over long training horizons. Finally, we find that existing mitigation techniques from the literature (along with our own) all reduce quantization error, suggesting a shared numerical mechanism behind their effectiveness. Together, our results identify finite precision error as a hidden driver of time-bomb spikes and show that suppressing this error is key to stable long-horizon pretraining.\blfootnote{Code available at: \href{https://github.com/tingtang2/quantization-loss-spikes}{https://github.com/tingtang2/quantization-loss-spikes}}

\end{abstract}

\section{Introduction}
\label{sec:intro}

Pretraining Transformer-based language models at scale is notoriously difficult due to the instability of the objective and its propensity to diverge or spike during training \citep{chowdhery-palm, zhang2022optopenpretrainedtransformer, zeng2023glmb, olmo20252olmo2furious}. Large-scale training
efforts have relied on heuristic procedures in order to recover from these spikes, including reverting to earlier checkpoints, skipping
data near observed spikes, and adjusting learning rates \citep{chowdhery-palm, almazrouei2023falconseriesopenlanguage, zhang2022optopenpretrainedtransformer}. While such interventions allow training to continue, they treat the symptoms of instability rather than its underlying cause.

A common class of loss spikes is associated with the growth and divergence of output logits in the language modeling (LM) head. \citet{chowdhery-palm} originally identified this behavior and introduced $z$-loss as a mitigation, an auxiliary loss penalty that discourages output logits from growing too large in magnitude. This mitigation has since become a standard safeguard in many training pipelines \citep{allal2025_the_smol_training_playbook_the_secrets_to_building_world_class_llms} and remains important in recent open-source, frontier-style training recipes. The latest Marin recipe includes $z$-loss alongside norm-preserving hyperball optimization, with leave-one-out ablations showing substantial degradation when $z$-loss is removed \citep{marin2026hero,marin2026loo}.

Despite the widespread adoption of methods for controlling output logit growth, our understanding of why such unmitigated growth leads to loss spikes remains incomplete. Recent work has traced the growth of logits to the optimizer, showing that Adam's adaptive preconditioning induces persistent drift in the mean output embedding of the LM head, which in turn causes the mean output logit to drift towards extreme values \citep{stollenwerk-stollenwerk-2025-better,stollenwerk2026outputembeddingcenteringstable}.  Yet, as we show in \Cref{sec:characterization_time_bomb}, this drift alone cannot substantively affect the forward or backward pass if we assume exact arithmetic. If this is the case, how can an apparently harmless drift ultimately destabilize training?

To unravel this mystery we focus our study on a particularly costly form of instability driven by output logit growth: slow, unrecoverable loss divergences that emerge only after prolonged training, which we call \textit{time-bomb spikes}. A run affected by a time-bomb spike can appear healthy for most of training before deteriorating irreversibly, after much of the compute budget has already been spent. While other kinds of fast, transient loss spikes exist, we find these are largely suppressed by learnable RMSNorm gains (\Cref{sec:characterization_fast}). Time-bomb spikes, in contrast, persist in these architectures and become more severe with longer training horizons, making them increasingly relevant as models are trained far beyond the compute-optimal token budget \citep{qwen3, everett2026optimizermemoryschedulesoutscaling}.

\textbf{In this work, we argue that the link between output logit growth and catastrophic loss spikes stems from how finite precision computation in the LM head leads to quantization error.} Using the time-bomb spike as a testbed, we explore simple interventions that act separately on each element of the puzzle, the drift itself and the precision of the LM head. First, we show that optimizing the LM head using SGD with momentum, rather than AdamW, eliminates these spikes by removing drift directly, indicating that the drift induced by preconditioning impacts stability in practice more than theory would predict. Second, we confirm that reducing quantization error by simply computing and optimizing just the LM head in higher precision is similarly sufficient for preventing time-bomb spikes. Finally, we tie together our LM head SGD-M mitigation, the effectiveness of increased precision, and other proposed methods in the literature by demonstrating that they all seem to work by suppressing quantization error through different mechanisms. In summary, our findings are:

\begin{enumerate}[topsep=0.0cm,itemsep=0.0cm,leftmargin=0.5cm,font=\bfseries]

  \item \textbf{LM head SGD-M effectively squashes spikes.}
    We propose replacing AdamW with carefully tuned SGD with momentum exclusively on the LM head.
    This targeted intervention suppresses logit growth and improves stability over long training horizons.

  \item \textbf{Finite precision is a hidden driver of time-bomb spikes.}
    Quantization error in the LM head grows rapidly before each failure, and loss diverges soon after. Precision directly controls when this happens. Lowering LM head precision triggers time-bomb spikes earlier, while computing the LM head in FP32 keeps runs stable over the horizons we study.

  \item \textbf{Existing mitigations all share a common numerical mechanism.}
    Even though our proposed LM head SGD-M and other mitigations from the literature intervene in different ways, they all keep quantization error bounded, whereas it grows by orders of magnitude in unmitigated runs. Suppressing quantization error thus emerges as a shared reason these seemingly unrelated methods prevent time-bomb spikes.
\end{enumerate}

\begin{figure}[t!]
  \centering
  \includegraphics[width=\columnwidth]{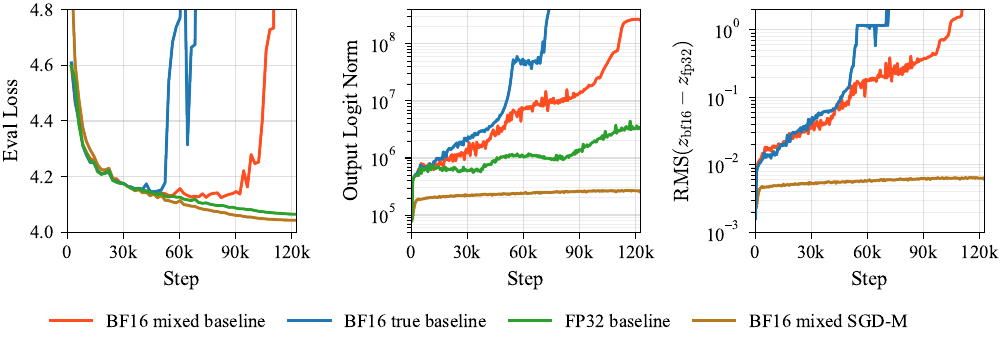}

  \caption{
    \textbf{Reduced precision accelerates time-bomb spikes, while LM head SGD-M prevents them.}
    Lowering LM head precision brings the failure earlier, with BF16 true diverging first, BF16 mixed later, and FP32 remaining stable throughout training (\textbf{left}). Output logit norm grows under every AdamW baseline, including FP32 (\textbf{middle}), but it destabilizes training only in reduced precision, where LM head quantization error grows by orders of magnitude before each failure (\textbf{right}). Quantization error is measured as the RMS difference between native BF16 logits and FP32 logits computed from the same hidden states. Replacing AdamW with SGD-M on the LM head keeps both logit norm and quantization error nearly flat, preventing the time bomb even in BF16 mixed precision. Results are for 28M parameter models trained at peak learning rate $0.1$ for ${\sim}28\times$ the Chinchilla optimal token horizon, using AdamW throughout (baselines) or SGD with momentum on the LM head (SGD-M).
  }
  \label{fig:fig_one}
\end{figure}

\section{Preliminaries}
\label{sec:preliminaries}

In this work, we investigate loss spikes stemming from output logit divergence in the language modeling head \citep{wortsman2024smallscale, chowdhery-palm}. This type of training instability was identified during the training of PaLM \citep{chowdhery-palm}, where the authors proposed $z$-loss as a mitigation technique.

\subsection{Language modeling head}
Consider a vocabulary of size $V$ and a final hidden
state $\bm{h} \in \mathbb{R}^d$. Let $\bm{y} \in \{0, 1\}^V$ denote the one-hot encoded ground truth labels for a single example such that $\sum_{i=1}^V y_i = 1$.
The output embedding matrix $\bm{E} \in \mathbb{R}^{V \times d}$ maps the final hidden state to output logits $\bm{z}=\bm{E}\bm{h} \in \mathbb{R}^{V}$. The
next-token distribution $\bm{p}$ and cross entropy objective are computed as
\begin{linenomath*}
  \[
    p_i
    = \operatorname{softmax}(\bm{z})_i
    = \frac{\exp(z_i)}{\sum_{j=1}^{V}\exp(z_j)},
    \quad i=1,\ldots,V.
    \qquad
    \mathcal{L}_{\mathrm{CE}}
    = -\sum_{k=1}^{V} y_k \log p_k
  \]
\end{linenomath*}

\subsection{Baseline Mitigations for Logit Divergence}

\textbf{$z$-loss.}
Output logit divergence occurs when the absolute value of the output logits $\bm{z}$ grows excessively large. This can lead to both fast spikes, where loss increases quickly but eventually recovers, as well as slow loss divergences as studied in \citet{wortsman2024smallscale}. $z$-loss penalizes the logits for drifting too far from zero by introducing a regularization term based on $Z = \sum_{j=1}^{V} e^{z_j}$ whereby the complete loss becomes $\mathcal{L} = \mathcal{L}_{CE} + \lambda \log^2{Z}.$

\textbf{$\mu$-centering.}
More recently, \citet{stollenwerk2026outputembeddingcenteringstable} introduced $\mu$-centering to precisely suppress output logit divergence. Unlike $z$-loss, which regularizes the loss objective, $\mu$-centering is a structural intervention as it centers the output embeddings around zero. Let $\bm{E} =
\begin{bmatrix} \bm{e}_1, \ldots, \bm{e}_V
\end{bmatrix}^\intercal$ be the output embedding matrix. After each optimization step, the mean output embedding $\bm{\mu}= \frac{1}{V}\sum_{i=1}^V \bm{e}_i$  is subtracted from each individual output embedding $e_i$, creating new output embeddings $\bm{e}_i^* = \bm{e}_i - \bm{\mu}.$ By enforcing $\sum_{i=1}^V \bm{e}_i^* = 0$, this procedure ensures the mean output logit remains zero, regardless of the hidden state magnitude.

\subsection{Experimental Setup}

To reproduce the loss spikes seen in larger models, we train 28M--1.4B parameter models at elevated learning rates following \citet{wortsman2024smallscale, zhao2025deconstructing, stollenwerk2026outputembeddingcenteringstable, su2025characterizationmitigationtraininginstabilities, li2025model} (fast and slow spike examples in \cref{fig:diff_lr,fig:time_bomb}). Models are implemented in JAX \citep{jax2018github} and Flax \citep{flax2020github} and, unless otherwise noted, trained on up to 32 Google Cloud v6e or v4 TPUs in bfloat16 mixed precision. We use a modernized GPT-2 \citep{radford2019language} architecture with rotary positional embeddings \citep{rope}, GELU activations \citep{hendrycks2023gaussianerrorlinearunits}, RMSNorm \citep{zhang-sennrich-neurips19}, untied embeddings, and no biases. We optimize with AdamW \citep{loshchilov2018decoupled} ($\epsilon = 10^{-8}$, $\beta_1 = 0.9$, $\beta_2 = 0.95$) using linear warmup, cosine decay, and gradient clipping at a global norm of $1$. Unless otherwise specified, weight decay is $10^{-4}$ and batch size is 262,144 tokens. For $z$-loss runs, $\lambda = 10^{-4}$ following \citet{chowdhery-palm}.

All runs use QK-Norm \citep{pmlr-v202-dehghani23a}, the recommended intervention against attention logit growth \citep{wortsman2024smallscale}. This removes that source of spikes, so we can study instabilities from output logit divergence in isolation. All models are trained on FineWeb \citep{penedo2024the} with the GPT-2 tokenizer and, unless otherwise noted, standard cross entropy loss.

\subsection{Warmup: Fast, Transient spikes}
\label{sec:characterization_fast}

We begin by briefly examining fast loss spikes, which
cause abrupt increases in loss before recovering. Unless otherwise stated, these experiments use parameter-less norms---RMSNorm layers without learnable gain parameters. This design choice is used in several large language model suites \citep{groeneveld-etal-2024-olmo, wang2026hrmtextefficientpretrainingscaling} and is also shared by prior work that studies or reports training instabilities
during language model pretraining \citep{geiping2025scaling,filatov2026optimalscalingneedsoptimal}.

We first investigate whether interventions targeting
individual spike-inducing tokens or batches can resolve
fast loss spikes. We find that masking as little as
0.2\% of high-loss tokens suppresses visible spikes
but degrades convergence, while skipping triggering
batches and reverting the optimizer state merely
postpones subsequent spikes
(\Cref{app:fast_spike_observation_results}).
These findings suggest that avoiding individual
spike-inducing examples is insufficient to eliminate
the model's susceptibility to fast spikes.  These spikes stem not only from data but from the parameters and/or optimizer state.

\begin{figure}[h!]
  \centering
  \includegraphics[width=\columnwidth]{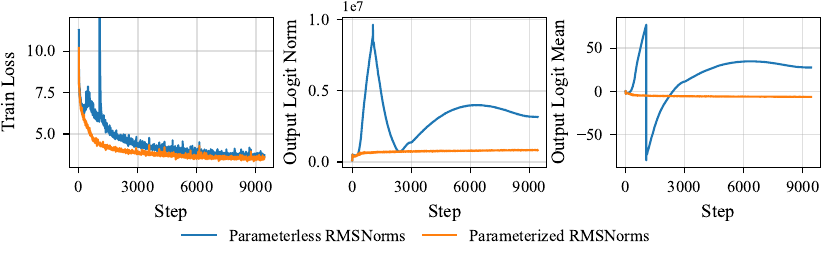}
  \caption{\textbf{Learnable gain parameters in normalization layers stabilize training and can prevent fast loss spikes.} Comparing a 162M spiking run with the same seed and configuration only differing in the inclusion of learnable RMSNorm parameters, the parameterized norm configuration exhibits no fast spikes, output logit swings, or large logit norm growth.}
  \label{fig:parameterized_vs_parameterless}
\end{figure}

\paragraph{Fast spikes depend strongly on normalization type.}
To test how strongly RMSNorm parameterization affects fast spikes, we repeat a representative fast spike configuration with learnable RMSNorm gains, holding all other hyperparameters and random seeds fixed. Across five matched seed pairs, parameterized RMSNorms substantially reduce the fast loss spikes and the associated output logit mean and norm growth. \Cref{fig:parameterized_vs_parameterless}
shows a representative matched pair, with all five comparisons reported in \Cref{fig:rmsnorm_companion_pairs}.

These results suggest that the abrupt fast spike regime studied above is strongly affected by RMSNorm parameterization and is substantially less prominent with the learnable gains used in many modern language models \citep{qwen3,megatron-lm}. Importantly, however, LM head instability does not disappear altogether. In the next section, we show that models with learnable RMSNorm gains can still develop a slower, delayed form of instability during long horizon training.

\section{Defusing Time-Bomb Loss Spikes}
\label{sec:characterization_time_bomb}

In this section, we turn our attention to a type of training instability that persists even when using parameterized norms: slow time-bomb spikes.
Experiments in this section use RMSNorm layers with learnable gain parameters. All models use a batch size of 131,072 tokens and no weight decay; all other training hyperparameters follow the shared setup in \Cref{sec:preliminaries}. We use this configuration because preliminary experiments indicated that smaller batch sizes lead to earlier time-bomb onset, while weight decay delays their emergence (as also noted in \citet{wortsman2024smallscale}). All models use bfloat16 mixed precision (BF16 Mixed) by default unless otherwise noted.

\subsection{Time-bomb loss spikes emerge during overtraining}
\label{sec:time_bomb_overtraining}

\begin{figure}[t!]
  \centering
  \includegraphics[width=\columnwidth]{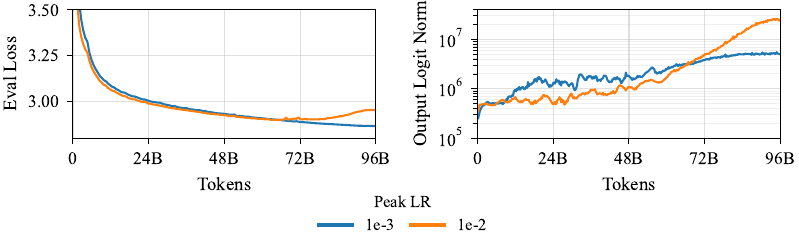}
  \caption{\textbf{Long horizon training can result in time-bomb instability.} Two 404M parameter models train similarly for most of pretraining, but the higher LR run exhibits delayed, unrecoverable loss increases accompanied by accelerating output logit growth.}
  \label{fig:time_bomb}
\end{figure}

To study time-bomb spikes, we overtrain 404M parameter models for 96B tokens across the same learning rate range as in \citet{wortsman2024smallscale}: $\{3\times10^{-4},\, 10^{-3},\, 3\times10^{-3},\, 10^{-2},\, 3\times10^{-2},\, 10^{-1},\, 3\times10^{-1}\}$. This corresponds to approximately 240 training tokens per parameter, or roughly \(12\times\) the Chinchilla-optimal token budget. Unless otherwise specified, these experiments are conducted on four NVIDIA B200 GPUs. \Cref{fig:time_bomb} shows two runs that follow nearly identical loss trajectories through most of training, but where the higher-learning rate run eventually stops improving and undergoes a sustained, monotonic increase in evaluation loss---a time-bomb.
Full loss and output logit norm curves for the learning rate sweep are shown in \cref{fig:full_time_bomb}. Longer training also leads to greater late stage output logit norm growth (Appendix \Cref{fig:token_horizon}), suggesting that time-bomb spikes become more of an issue in the overtraining regime.

\subsection{Adam's preconditioner induces LM head mean drift}
\label{sec:lm_head_gradients}
Prior work has shown that the cross entropy gradient of the LM head has a zero-sum structure along the
vocabulary axis, while Adam's coordinate-wise preconditioning breaks this property \citep{bis-etal-2021-much,stollenwerk-stollenwerk-2025-better}.
This can induce persistent drift in the mean output embedding and,
by extension, in the mean output logits, directly linking the logit growth dynamics
observed in the previous experiments to an interaction between the LM head gradients and Adam's
preconditioner \citep{stollenwerk-stollenwerk-2025-better}.

Let $\bm{E}\in\mathbb{R}^{V\times d}$ denote the output embedding matrix
and $\bm{h}\in\mathbb{R}^d$ the incoming hidden activations, such that
$\bm{z}=\bm{E}\bm{h}$. The cross entropy gradient admits the following zero-sum property:
\[
  \frac{\partial \mathcal{L}_{CE}}{\partial \bm{E}}
  =(\bm{p}-\bm{y})\bm{h}^{\intercal},
  \quad\quad
  \sum_{i=1}^{V} \frac{\partial \mathcal{L}_{CE}}{\partial E_{i,j}}
  = h_j \underbrace{\sum_{i=1}^{V} (p_i - y_i)}_{=\,0} = 0,
\]
because softmax probabilities and one-hot labels both independently sum to one. Thus, any optimizer update that preserves this zero-sum structure leaves the mean output embedding over the vocabulary axis
$\bm{\mu}=\frac{1}{V}\sum_i\bm{e}_i$ unchanged. Although Adam's first moment estimate remains zero-sum, division by
the coordinate-dependent second moment reweights vocabulary rows
non-uniformly. Because these second moments vary strongly with token
frequency \citep{stollenwerk-stollenwerk-2025-better}, the resulting Adam
update can have a non-zero vocabulary mean and induce drift in
$\bm{\mu}$.

\paragraph{Larger learning rates and longer training
horizons amplify logit growth.}
Because Adam's preconditioned updates need not have zero mean across each word embedding along the vocabulary axis, larger learning rates can amplify their contribution to the mean output embedding at each optimization step. Over longer training horizons, these updates have more opportunities to accumulate. This provides a possible explanation for the greater output logit growth observed at higher learning rates (\Cref{fig:full_time_bomb}) and over longer training horizons (\Cref{fig:token_horizon}).

\subsection{Mitigating Drift in the LM Head}
\label{sec:mitigation_methods}

The zero-sum structure motivates a simple optimizer level intervention: train the LM head with SGD with momentum while retaining AdamW for the remainder of the model. Unlike Adam, SGD with momentum (SGD-M) does not apply coordinate-wise adaptive rescaling and therefore preserves the zero-sum structure of the LM head gradient. We compare this approach against AdamW, $\mu$-centering, and $z$-loss.

\begin{figure}[t]
  \centering
  \includegraphics[width=\columnwidth]{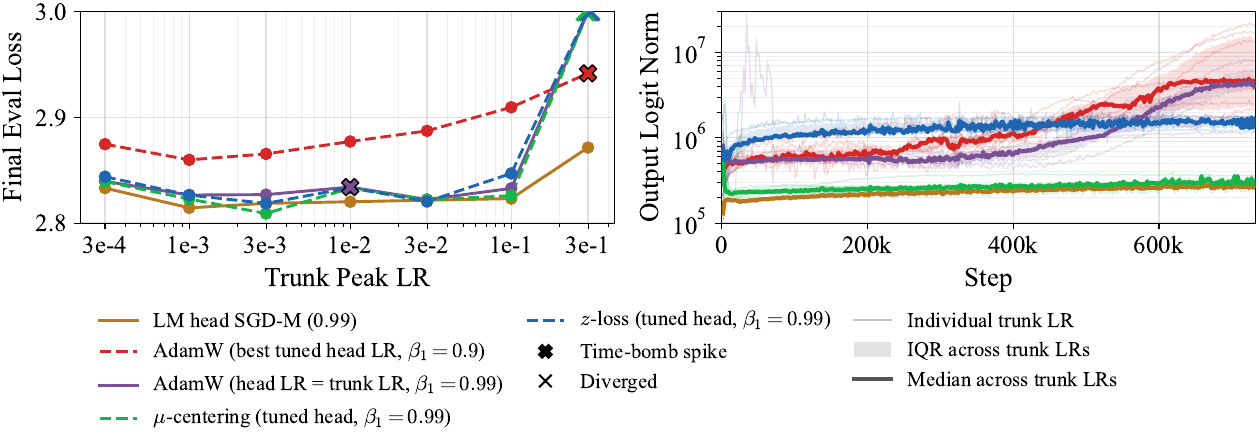}
  \caption{\textbf{Increasing LM head $\beta_1$ is more effective than tuning its learning rate alone.}
    \textbf{Left:} Final evaluation loss across trunk learning rates. Head learning rate tuning yields modest gains, while $\beta_1=0.99$ closes much of the gap to LM head SGD-M. Adding $\mu$-centering or $z$-loss improves some settings, but all AdamW variants diverge or time-bomb at the largest trunk learning rate, whereas SGD-M remains stable. Dashed lines denote tuned head learning rates.
  \textbf{Right:} output logit norm trajectories, showing how each method constrains LM head growth.}
  \label{fig:mitigations_tuned_400M}
\end{figure}

\paragraph{Comparing mitigation methods.}
For LM head SGD-M, we tune the head learning rate separately and use increased momentum of 0.99. To ensure a fair comparison, we also allocate the same hyperparameter tuning budget to the AdamW based methods, as discussed below and shown in \Cref{fig:mitigations_tuned_400M}.

\begin{figure}
  \centering
  \includegraphics[width=\columnwidth]{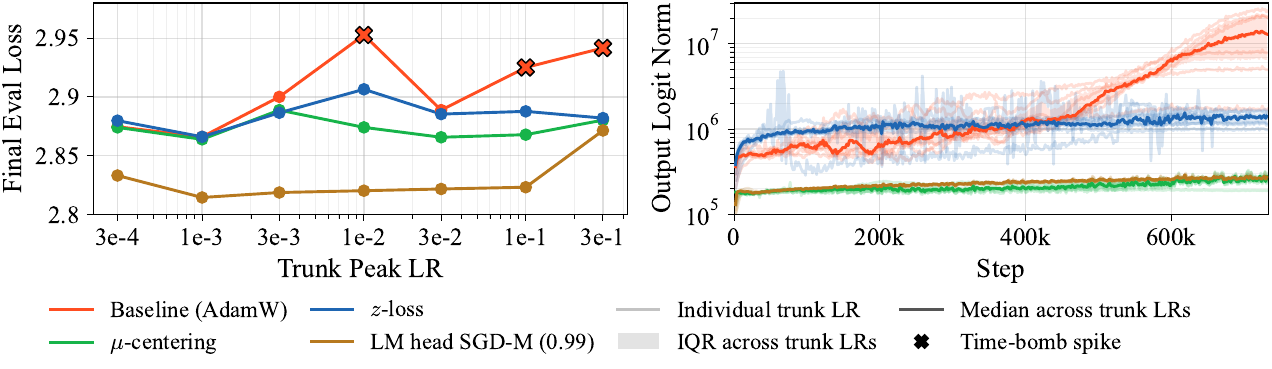}
  \caption{\textbf{SGD-M is competitive in both optimization quality and control of LM head growth.} Across trunk learning rates, \(\mu\)-centering, \(z\)-loss, and LM head SGD-M improve stability relative to the AdamW baseline, while differing in both final evaluation loss (left) and the degree of LM head growth suppression (right). Across high learning rates, $\mu$-centering and LM head SGD-M strongly suppress output logit norm growth, whereas the baseline develops the largest late growth and corresponding time-bomb spikes (right). Faint curves show individual learning rates; solid curves and shaded regions show the median and interquartile range across learning rates.}
  \label{fig:mitigations_400M}
\end{figure}

We first compare the methods without head-specific
retuning of the AdamW variants in
\Cref{fig:mitigations_400M}, which shows results across
the 404M-parameter learning rate sweep. All three methods improve stability relative to the AdamW baseline and substantially reduce output logit growth. In particular, LM head SGD-M strongly constrains logit norm growth while remaining competitive in final evaluation loss across the sweep. In Appendix \cref{sec:fast_spikes_mitigation_results}, we also compare these mitigations in the parameter-less norm, fast spikes regime from \cref{sec:characterization_fast} and find that they are similarly effective against this class of instability.

\paragraph{LM head specific tuning.}
\label{para:lm_head_tuning}
For LM head SGD-M, the highest head learning rate in our grid achieves the best performance across all trunk learning rates and extending the search to larger values yields no meaningful further improvement. In order to measure whether head specific hyperparameter tuning gives SGD-M a disproportionate advantage, in \Cref{app:lm_head_hyperparam} and \cref{fig:mitigations_tuned_400M} we explore the results of tuning the head LR and head specific $\beta_1$ value for AdamW. Tuning the AdamW head-specific learning rate alone provides only modest gains, while increasing the first moment coefficient for the LM head to \(\beta_1=0.99\) closes much of the optimization gap between AdamW and SGD-M. This is consistent with recent work showing that longer training horizons can require retuning optimizer hyperparameters across the overtraining axis \citep{everett2026optimizermemoryschedulesoutscaling}. We also explore combining the higher-\(\beta_1\) AdamW head with \(\mu\)-centering and \(z\)-loss. However, despite their improved performance over the tuned AdamW at some learning rate settings, those combinations both also diverge at the largest trunk learning rate we test.

\FloatBarrier
\begin{summarybox}[SGD-M on the LM head is an effective stabilizer]
  Replacing AdamW with SGD with momentum
  exclusively on the LM head suppresses logit growth
  and improves long-horizon training stability.
  Increasing AdamW's first moment coefficient recovers much of the optimization gap, but does not consistently prevent time-bomb instability.
\end{summarybox}

\subsection{Finite Precision's Role in Loss Spikes}

\paragraph{Mean preservation alone cannot explain training stabilization.}
Although the analysis and experiments above connect Adam's preconditioner to drift in the vocabulary-mean output embedding, controlling this component alone cannot
affect the cross entropy computation in exact arithmetic.

To see why, let $\bm{E}'=\bm{E}-\bm{1}\bm{\mu}^{\intercal}.$
For a fixed hidden state, this transformation shifts every logit by the
same scalar and therefore leaves the softmax distribution unchanged
\citep{stollenwerk2026outputembeddingcenteringstable}. Because $\bm{p}$ is
unchanged, the cross entropy gradient with respect to the LM head is consequently unchanged as well. Centering also leaves the gradient propagated into the hidden state unchanged:
\[
  \begin{aligned}
    \frac{\partial \mathcal{L}_{CE}}{\partial \bm{h}}
    &= \bm{E}'^{\intercal}(\bm{p}-\bm{y}) = \bm{E}^{\intercal}(\bm{p}-\bm{y})
    -\bm{\mu}\sum_{i=1}^{V}(p_i-y_i) = \bm{E}^{\intercal}(\bm{p}-\bm{y}).
  \end{aligned}
\]
Therefore, in settings without weight decay, and assuming exact computation, removing the vocabulary-mean component changes neither the model predictions nor the cross entropy gradients through the LM head. Thus, in theory, growth in this mean component alone should not affect the model's predictions or optimization trajectory, nor should the centering performed by interventions like $\mu$-centering meaningfully stabilize training.

However, in \textit{finite} precision, a large logit mean implies an increase in the absolute numerical scale of the individual logit values, potentially reducing the precision with which the differences between logits that determine the predictive distribution can be represented. In our final set of experiments, we explore how the effect of limited precision in the LM head can help explain both the sensitivity to logit mean drift that real language models exhibit as well as the apparent effectiveness of mitigations like $\mu$-centering and $z$-loss by accounting for changes in arithmetic representation error.

\paragraph{Reduced precision in the LM head accelerates time-bombs.}\label{sec:precision_accelerates_time_bomb}

\begin{figure}
  \centering
  \includegraphics[width=\columnwidth]{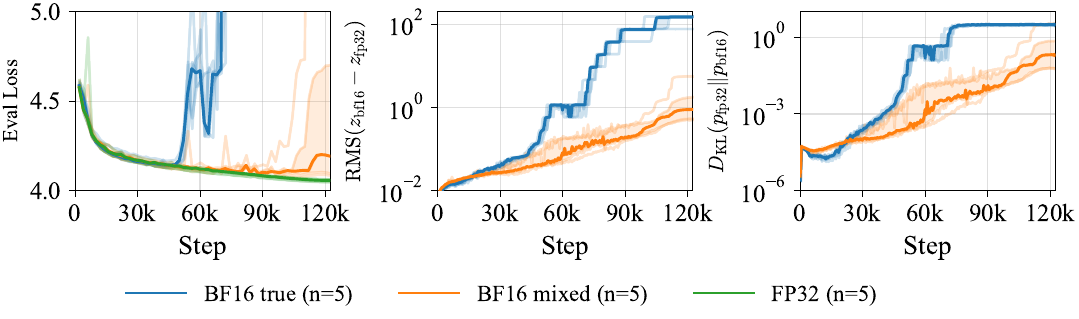}
  \caption{\textbf{Reduced precision in the LM head accelerates time-bomb instability and increases quantization error as logits grow.} For 28M parameter models trained at peak LR \(0.1\), BF16 true runs fail earliest, BF16 mixed runs deteriorate later, and FP32 remains stable across five seeds (left). The same ordering appears in the quantization error between BF16 native and FP32 reference logits (middle) and in the KL divergence between the corresponding predictive distributions (right), indicating that loss spikes coincide with increasingly large finite precision error. Solid lines show the median across seeds and faint lines show individual runs.}
  \label{fig:precision_ablation}
\end{figure}
For these experiments, we overtrain 28M parameter models for 16B tokens using the same learning rate range as before. These models are thus training for approximately 570 tokens per parameter, or roughly 28$\times$ the Chinchilla-optimal token budget. We vary LM head precision across three settings: FP32, BF16 mixed, and BF16 true, ordered from highest to lowest numerical precision. BF16 mixed, the standard setting for large-scale pretraining, stores FP32 master weights but casts to BF16 at runtime for faster matrix multiplications and lower activation memory; BF16 true additionally stores the LM head weights and optimizer states in BF16 (see \cref{app:precision_settings} for details).

To directly measure numerical error in the LM head, at each logging step we recompute the LM head logits in FP32 from the same hidden states used by the native forward pass, yielding FP32 reference logits \(z_{\mathrm{fp32}}\). For these reference logits, the LM head GEMM is also performed in FP32. We compare these reference logits against the native BF16 logits \(z_{\mathrm{bf16}}\), measuring logit quantization error as the RMS distance, $\mathrm{RMS}(z_{\mathrm{bf16}}-z_{\mathrm{fp32}})$, analogous to the commonly used reconstruction error in quantization literature \citep{frantar2023optq}. We also measure the
predictive-distribution quantization error as the KL divergence between the corresponding post-softmax distributions, $D_{\mathrm{KL}}(p_{\mathrm{fp32}}\|p_{\mathrm{bf16}})$, following \citet{lotfi2026quantizedreasoningmodelsthink}.

\Cref{fig:precision_ablation} shows that BF16 true runs develop substantial numerical error and time-bomb first, BF16 mixed runs fail a bit later, but FP32 runs remain stable over the training horizon. In both reduced-precision settings, increasing logit quantization error is accompanied by increasing predictive-distribution quantization error and an eventual loss spike. We present a deeper analysis of the precision-error correspondence in \cref{app:precision_analysis}.

\begin{summarybox}[Finite precision drives time-bomb spikes]
  Although mean output logit drift is harmless in exact arithmetic, it amplifies numerical error under finite precision. Reducing LM head precision accelerates time-bomb spikes, while increasing precision improves stability.
\end{summarybox}

\paragraph{All mitigation methods suppress quantization error.}

\begin{figure}
  \centering
  \includegraphics[width=\columnwidth]{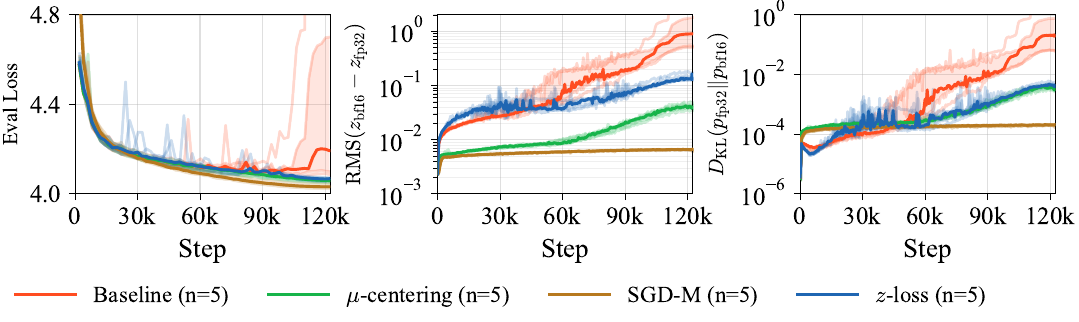}
  \caption{\textbf{Methods that prevent time-bomb failure suppress quantization error.} For 28M-parameter models trained in BF16 mixed at peak LR \(0.1\), the unmitigated baseline develops time-bomb spikes (left) together with orders-of-magnitude growth in logit quantization error (middle) and predictive distribution quantization error (right). All mitigation methods prevent the time bomb and keep both numerical-error metrics bounded across five seeds. Solid lines show medians and faint lines show individual runs. The same effect is more pronounced under BF16 true training; the unmitigated baseline fails catastrophically after significant growth in numerical error (\Cref{fig:mitigations_30M}).}
  \label{fig:mitigations_28M_bf16_mixed}
\end{figure}

Finally, we test the hypothesis that different mitigation methods like $z$-loss and $\mu$-centering may work at least in part because they suppress the growth in numerical errors associated
with time-bomb instabilities. \Cref{fig:mitigations_28M_bf16_mixed}
evaluates some of the methods from the previous section under BF16 mixed training. We observe that while the unmitigated baseline exhibits time-bomb loss spikes accompanied by a significant rise in quantization error in both logits and predictive distribution, in contrast,
$\mu$-centering, $z$-loss, and LM head SGD-M all prevent time-bomb failure and
keep both error metrics bounded across five seeds. Thus, methods that stabilize
LM head growth and prevent time-bomb spikes also suppress runaway growth in finite precision error, suggesting an underlying mechanistic connection between these independently proposed approaches.

\begin{summarybox}[All roads lead to quantization error control]
  $z$-loss, $\mu$-centering, and LM head SGD-M act on the loss objective, the output embeddings, and the optimizer, respectively. Yet all three prevent time-bomb spikes and keep quantization error bounded, while error in the unmitigated baseline grows by orders of magnitude. Controlling finite precision error thus emerges as a common thread behind these otherwise distinct methods.
\end{summarybox}

\paragraph{Practical considerations.}
The interventions studied differ significantly in their computational and memory costs. Computing the LM head in FP32 is the most direct fix but also the most expensive. FP32 GEMMs run at a fraction of BF16 throughput on modern accelerators, and the LM head is one of the largest matrix multiplications in the model, with the output dimension equal to the vocabulary size. Materializing FP32 logits also doubles the memory of what is often the largest activation tensor in training, a cost that grows with batch size, sequence length, and vocabulary size. By contrast, the other mitigations retain BF16 mixed precision and add little overhead. LM head SGD-M replaces AdamW only for the output embeddings and stores one fewer optimizer moment buffer, reducing memory; $z$-loss adds an auxiliary loss term; and $\mu$-centering requires a reduction across vocabulary rows after each update. These methods therefore keep the LM head in a regime where reduced precision suffices, without paying the cost of higher precision.

\section{Related Work}

\paragraph{Loss spikes in language model pretraining.}
Loss spikes have been widely documented in large-scale language model pretraining \citep{chowdhery-palm, zhang2022optopenpretrainedtransformer, workshop2022bloom, zeng2023glmb, deepseekai2026deepseekv4, kimiteam2026kimik2openagentic}, motivating a variety of empirical stabilization strategies. These include skipping problematic batches or rewinding to earlier checkpoints \citep{chowdhery-palm, zhang2022optopenpretrainedtransformer}, modifying gradient clipping or optimizer states \citep{pascanu2013difficultytrainingrecurrentneural, huang2025spam}, and architectural interventions targeting embedding or attention-layer instabilities \citep{zeng2023glmb, workshop2022bloom, dettmers20218, henry2020query, pmlr-v202-dehghani23a, kimiteam2026kimik2openagentic}. While these approaches can improve training stability, they address different potential sources of instability without necessarily identifying the mechanisms actually responsible for spikes. Notably, \citet{chowdhery-palm} found that replaying the same batches from different checkpoints did not reproduce loss spikes, suggesting that problematic training examples alone cannot explain their occurrence. We extend this observation by distinguishing the immediate triggers of loss spikes from the underlying instability that makes models susceptible to them. We find that fast spikes are highly sensitive to individual tokens and their contexts, yet skipping spike-inducing batches merely postpones subsequent failures a small number of iterations. We further uncover the cause of slow, terminal ``time-bomb'' spikes that emerge only after prolonged training, even in architectures where fast spikes are largely suppressed. This regime is likely to become more relevant as models are trained further beyond compute-optimal budgets.

\paragraph{Optimization dynamics of the LM head.}
A growing body of work identifies the LM head as particularly sensitive to the interaction between language modeling objectives and adaptive optimization. The highly imbalanced token frequencies in language modeling induce heterogeneous gradient statistics across vocabulary rows, and Adam's coordinate-wise adaptivity has been shown to play an important role in optimizing these parameters \citep{kunstner2024heavy, zhao2025deconstructing}. However, this same adaptivity can introduce undesirable behavior in the output embeddings. \citet{bis-etal-2021-much} study common shifts in Transformer embeddings, while \citet{stollenwerk-stollenwerk-2025-better} connect output embedding drift to Adam's preconditioner, which breaks the zero-sum structure of the cross entropy gradient along the vocabulary axis. This creates an apparent tension: the adaptivity that benefits LM head optimization can also induce persistent growth in the mean output embedding. Our work builds on this observation by examining the stability consequences of this drift over long training horizons. Crucially, because shifting all output logits by a common value leaves the softmax distribution and cross entropy gradients unchanged in exact arithmetic, mean drift alone cannot explain the resulting instability. We show that finite precision computation provides a mechanism connecting this otherwise harmless growth to training failure.

\paragraph{Finite precision and training instability.}
Concurrent work has identified several ways that numerical errors in softmax gradients can destabilize training. \citet{chen2026brokensymmetrybf16attention} show that BF16 rounding in the backward pass of fused attention kernels can violate the zero-sum structure of attention softmax gradients, causing delayed training instability. In contrast, our work focuses on quantization errors in the forward computation of the LM head caused by logit drift induced by Adam's preconditioner.

\citet{liu2026grokkingglitchinglowprecisiondrives} and \citet{barron2026fixsoftmaxcrossentropy} show that numerical errors in softmax cross entropy gradients underlie the Slingshot phenomenon \citep{thilak2022slingshotmechanismempiricalstudy}, a pattern of abrupt, recoverable loss spikes observed primarily in grokking rather than language modeling. \citet{barron2026fixsoftmaxcrossentropy} evaluate on synthetic modular arithmetic tasks, while \citet{liu2026grokkingglitchinglowprecisiondrives} also examine language modeling, finding abnormal logit growth during GPT training but no Slingshot spikes. We instead study slow, unrecoverable time-bomb spikes in the LM head, which arise when Adam induced logit drift, harmless in exact arithmetic, is amplified by quantization error in the forward logit computation. Consistent with this distinction, replacing standard cross entropy with the fixed softmax of \citet{barron2026fixsoftmaxcrossentropy} does not prevent the underlying logit growth and quantization error, and delays but does not prevent time-bomb spikes under BF16 true (\Cref{app:softmax_fix_results}).

\paragraph{Mitigating output logit divergence.}
Several methods have been proposed to stabilize training by constraining output logit growth, although they intervene at different points in the optimization process. The $z$-loss introduced by \citet{chowdhery-palm} penalizes large log-partition values and has become a common component of language model training recipes \citep{allal2025_the_smol_training_playbook_the_secrets_to_building_world_class_llms, marin2026hero, marin2026loo}. Similarly, Gemma 2 \citep{gemmateam2024gemma2improvingopen}
employs logit soft-capping, applying a scaled $\tanh$
transformation to directly bound the output logits
before the softmax operation.

Other approaches instead intervene on the output
embeddings or their optimization dynamics.
\citet{stollenwerk2026outputembeddingcenteringstable}
propose $\mu$-centering, which explicitly removes
the vocabulary-mean component of the output embedding
after each optimization step, while
\citet{stollenwerk-stollenwerk-2025-better} propose
modifying Adam's second-moment preconditioning to
prevent this drift from arising.
These approaches constrain different aspects of
LM head growth, ranging from directly bounding logits
to modifying the objective, output embeddings,
or optimizer updates.

Motivated by the zero-sum structure of the LM head
gradient, we revisit the use of SGD with momentum
specifically on the LM head, which preserves this
structure without requiring an explicit centering
operation. Beyond comparing the effectiveness of
the interventions we evaluate, we investigate why
they improve stability despite acting through
different mechanisms. Our experiments show that
$z$-loss, $\mu$-centering, and LM head SGD-M all
suppress the growth of quantization error in the
LM head, and that higher precision LM heads do not require these interventions, suggesting that controlling finite precision
error is an important shared mechanism underlying
their effectiveness against time-bomb spikes.

\section{Conclusion}

We study time-bomb spikes, a deleterious form of loss divergence that emerges in standard language model architectures after prolonged pretraining. Prior work traces the underlying logit growth to Adam's preconditioner, which breaks the zero-sum structure of the cross entropy gradient and causes the mean output embedding to drift. Yet this drift leaves both the forward and backward pass unchanged in exact arithmetic. We show that finite precision resolves this puzzle. As the logit mean grows, quantization error in the LM head grows by orders of magnitude before each failure, and the precision of the LM head computation directly controls when failures occur. Lowering precision brings time-bomb spikes earlier, while computing the LM head in FP32 keeps runs stable over the horizons we study.

This perspective also clarifies why our proposed application of SGD-M to the LM head and existing mitigations seem to work so well. Replacing AdamW with carefully tuned SGD with momentum on just the LM head prevents mean drift at its source and stabilizes training across learning rates where unmitigated training runs eventually diverge. Similarly, while $z$-loss, $\mu$-centering, and LM head SGD-M intervene at different points in the optimization process, we show that all approaches keep quantization error bounded. Rather than competing explanations of instability, these methods can be understood as different ways of keeping the LM head within a regime where finite precision suffices.

\textbf{Future Work.} We simulate spiking settings using higher than normal learning rates (as is common in \citet{zhao2025deconstructing, stollenwerk2026outputembeddingcenteringstable, su2025characterizationmitigationtraininginstabilities, li2025model}) but
we acknowledge that our findings may not fully transfer to spikes that arise in more natural settings in larger models, and it remains to be seen whether LM head SGD-M similarly outperforms standard optimizers there. Other avenues for future work include building specialized optimizers for the LM head and evaluating these mitigations in even lower precision formats such as FP8.

\subsection*{AI use statement}

We used generative AI tools (Gemini, ChatGPT, Claude, Codex, and ARIA) at several stages of this work. During research, they assisted with brainstorming, discovering research topics and identifying gaps, refining hypotheses, suggesting experimental parameters, providing feedback on experiments, interpreting results, and supporting qualitative and thematic data analysis. During implementation, they assisted with implementing methods and with creating and editing software code. During writing, they assisted with creating scientific figures and editing the paper to improve readability.

We did not use generative AI tools for cleaning or formatting datasets, helping develop theoretical models, or formulating mathematical claims. Providing critical ingredients for proving mathematical claims, assisting in the writing of proofs, generating synthetic datasets, and translation are not applicable to this work.

All experiments were designed and run by the authors, and all reported results come from those experiments. We reviewed all AI-assisted work. LLM-generated code was verified and tested for correctness, and LLM-retrieved papers were checked to confirm that they are relevant and correctly cited. The authors take full responsibility for the final content of this work, including any text, claims, or artifacts produced with the aid of generative AI.

\section*{Acknowledgments}
We gratefully acknowledge use of the research computing resources of the Empire AI Consortium, Inc \citep{Bloom2025EmpireAI}, with support from Empire State Development of the State of New York, the Simons Foundation, and the Secunda Family Foundation. This work was also supported by Google’s \href{https://sites.research.google/trc/about/}{TPU Research Cloud (TRC) program}. MG and ZC received generous support from the Columbia University Stabilization Award and from the Center For AI and Responsible Financial Innovation.

\raggedbottom
\bibliographystyle{iclr2027_conference}
\bibliography{paper_refs}

\newpage
\flushbottom
\appendix
\section{Additional fast spike results}

\subsection{Learning rate controls the type of instability}

\begin{figure}[H]
  \centering
  \includegraphics[width=0.7\columnwidth]{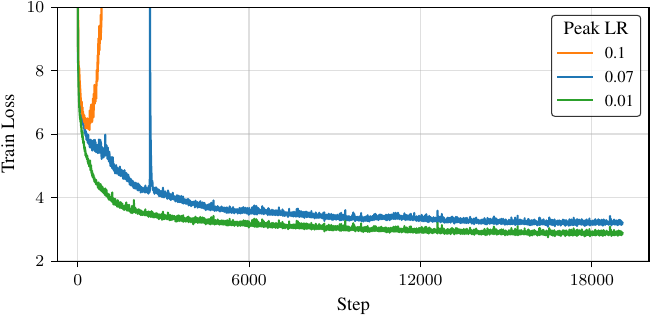}
  \caption{\textbf{Learning rates can control the type of training instability elicited}. Example training runs at 1.4B scale trained for 10B tokens. We focus on the learning rates similar to the middle one in order to elicit fast loss spikes to analyze.}
  \label{fig:diff_lr}
\end{figure}

\FloatBarrier
\subsection{Fast spikes concentrate in outlier tokens}
\label{app:fast_spike_observation_results}

\begin{figure}[H]
  \centering
  \includegraphics[width=\columnwidth]{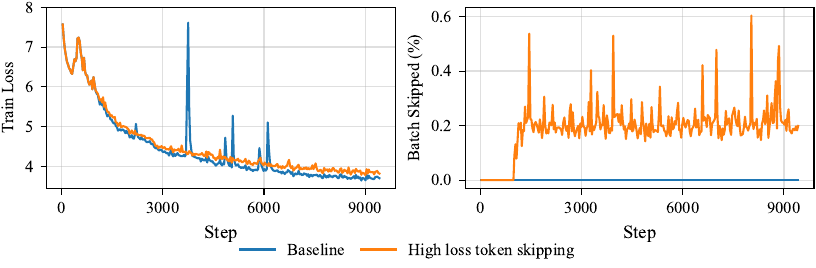}
  \caption{\textbf{Skipping a small fraction of high-loss tokens eliminates loss spikes but hurts convergence.} A 162M model trained at the Chinchilla-optimal token budget with peak learning rate 0.15. One run is the baseline; the other skips high-loss tokens identified via a rolling window threshold.
  }
  \label{fig:loss_skipping}
\end{figure}

During a fast loss spike, we find that a small subset of tokens incur extremely large cross entropy losses (100+) and therefore are responsible for the majority of the spike's apparent magnitude. To quantify these outliers, we compute a robust $z$-score using a rolling window of token losses from the preceding 256 steps (after an initial warmup period at the beginning of the run). This is similar in spirit to \citet{karpathy2024tweet}, but uses a robust $z$-score as we find token loss distributions to be heavy-tailed. As shown in \Cref{fig:loss_skipping}, masking as little as 0.2\% of tokens is sufficient to eliminate major spikes. This masking, however, consistently hurts overall convergence in training loss.

Upon observing that skipping tokens can be effective at eliminating fast spikes, one might conclude that loss spikes stem from exposure to bad training examples and could be eliminated by filtering the data. We interrogate this hypothesis further in \cref{app:swaps} by identifying extreme outlier tokens and perturbing their contexts to gain a more mechanistic understanding of the role they play in causing a spike. While these token-level interventions are also effective at reducing the loss during a spike, the token types and context-sensitive perturbations required to achieve stability in this manner appear to be non-systematic and therefore do not admit a practical train-time strategy that one could deploy during a real training run.

\FloatBarrier

\subsection{Spike events are highly sensitive to context}
\label{app:swaps}
\raggedbottom

To understand what happens at the token level during a fast spike, we rank validation-set predictions by cross entropy loss at the peak of a spike in the 162M run from \Cref{fig:logit_flip}, aggregating across all batches, sequences, and positions. We refer to the highest-loss prediction as the \textbf{Rank-1 event}. In this event, the model receives the context ``\texttt{...the commencement of the 50th Commemoration of that War.The Command}'' and must predict the target token \texttt{ery}, which completes the word ``Command\textbf{ery}''. The model instead assigns a logit of $782.05$ to the unrelated token \texttt{ible}, incurring a loss of $610.0$ nats (\Cref{tab:logit_swap_analysis}).

\begin{figure}[H]
  \centering
  \includegraphics[width=\linewidth]{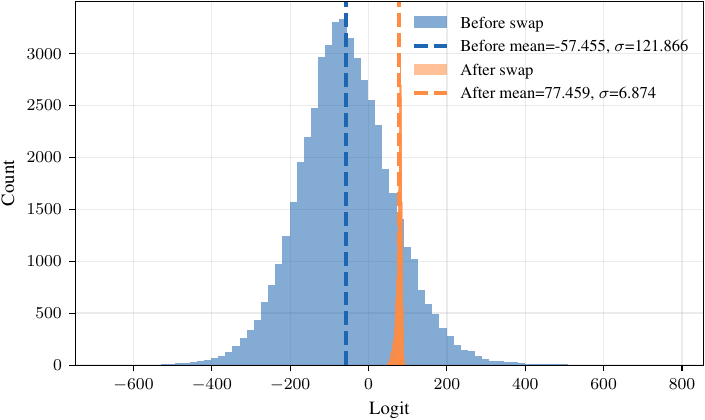}
  \caption{\textbf{A single-token context swap collapses the logit distribution at a spike.} Full-vocabulary logits ($V = 50{,}260$) at the Rank-1 spike event before (blue) and after (orange) replacing the final context token \texttt{Command} with a comma, with the target token \texttt{ery} held fixed. The standard deviation contracts $17\times$, from $121.87$ to $6.87$, recovering the range typical of non-spiking steps.}
  \label{fig:logit_histogram_rank1}
\end{figure}

This spike is remarkably fragile. Replacing only the final context token, \texttt{Command}, with a comma while keeping the target fixed reduces the loss from $610.0$ to $12.7$ nats, and the largest logit falls from $782.05$ to $95.01$. \Cref{fig:logit_histogram_rank1} shows that the swap significantly changes the logit distribution. Before the swap, the logits have a standard deviation of $121.87$, so even low-probability tokens receive extreme values. After the swap, the standard deviation shrinks to $6.87$, which is typical of the model's behavior during non-spiking steps.

We observe similar recoveries when swapping tokens at other positions in the context and when using other replacement tokens, including the single space \texttt{' '} and empty space \texttt{''} tokens. Spike events are therefore highly sensitive to context. Small perturbations to the preceding tokens are often enough to restore normal loss values, and the spike itself corresponds to the model placing nearly all of its probability mass on a single wrong token.

\begin{table}[H]
  \centering
  \begin{tabular}{l l r l r}
    \toprule
    & \multicolumn{2}{c}{\textbf{Original context (spiking)}} & \multicolumn{2}{c}{\textbf{Swapped context (recovered)}} \\
    \cmidrule(lr){2-3} \cmidrule(lr){4-5}
    \textbf{Rank} & \textbf{Token} & \textbf{Logit} & \textbf{Token} & \textbf{Logit} \\
    \midrule
    1 & \texttt{ible}  & 782.05 & \texttt{\textvisiblespace the}     & 95.01 \\
    2 & \texttt{ments} & 669.86 & \texttt{\textvisiblespace in}      & 94.12 \\
    3 & \texttt{ured}  & 640.67 & \texttt{\textvisiblespace however} & 93.80 \\
    4 & \texttt{ive}   & 603.56 & \texttt{\textvisiblespace until}   & 92.99 \\
    5 & \texttt{ivism} & 595.26 & \texttt{\textvisiblespace and}     & 92.96 \\
    \midrule
    \textbf{Target} & \texttt{ery} & --- & \texttt{ery} & --- \\
    \textbf{Loss}   &              & \textbf{610.00} & & \textbf{12.73} \\
    \bottomrule
  \end{tabular}
  \caption{\textbf{A single-token context swap collapses the spike.} Top-5 predicted tokens at the Rank-1 spike event before and after replacing the final context token \texttt{Command} with a comma. Before the swap, the model assigns extreme logits to unrelated subword tokens; after the swap, the top predictions are ordinary function words with logits in a normal range, and the loss falls from $610.00$ to $12.73$. \texttt{\textvisiblespace} denotes a leading space.}
  \label{tab:logit_swap_analysis}
\end{table}

\newpage
\flushbottom

\FloatBarrier
\subsection{Skipping triggering batches only delays spikes}
\label{app:logit_swings}

\begin{figure}[htbp]
  \centering
  \includegraphics[width=\columnwidth]{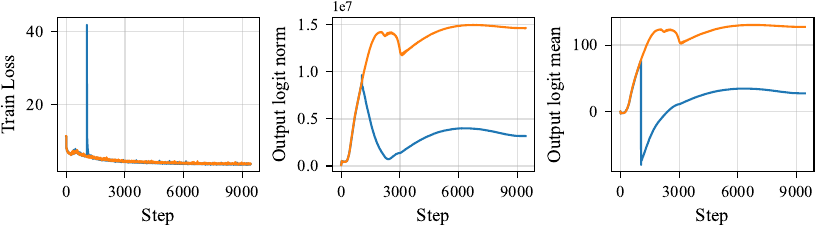}
  \caption{\textbf{Fast loss spikes coincide with sharp swings in the output logit mean.}
    A spiking 162M run compared with a non-spiking run using the
  same configuration and differing only in random seed.}
  \label{fig:logit_flip}
\end{figure}

\paragraph{Output logit mean dynamics during fast spikes.}
Fast spikes coincide with sharp swings in the mean output logit. As shown in
\Cref{fig:logit_flip}, these changes occur alongside the onset of a spike and
are absent in an otherwise matched non-spiking run differing only in random
seed.

\paragraph{Removing triggering batches only delays fast spikes.}

\begin{figure}[htbp]
  \centering
  \includegraphics[width=\columnwidth]{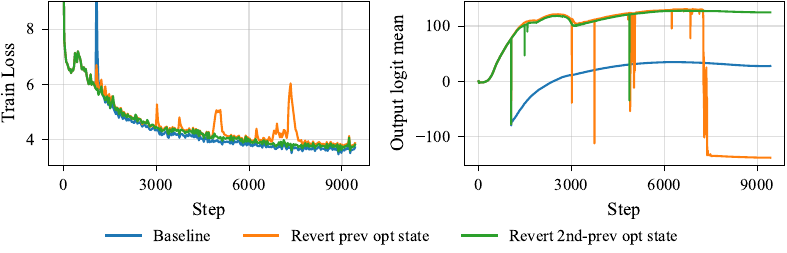}
  \caption{\textbf{Skipping high-loss batches defers rather than eliminates
    fast spike instability.} Following \citet{karpathy2024tweet}, we skip any batch whose loss exceeds 7 standard
    deviations from a rolling 1000-step window and revert the optimizer state.
    Reverting to either the previous or second-to-previous optimizer state does
  not prevent subsequent fluctuations in output logit mean characteristic of spike events.}
  \label{fig:batch_skipping}
\end{figure}

We also observe that fast loss spikes frequently coincide with sharp changes in the mean output logit, as shown in \Cref{fig:logit_flip}. We therefore ask whether these events can be treated in the same way as the high-loss outlier tokens by skipping the \emph{batch} that induces a loss spike. Following the batch-skipping procedure of \citet{karpathy2024tweet}, we find that skipping high-loss batches merely defers the instability rather than resolving it. As shown in \Cref{fig:batch_skipping}, after reverting to an earlier optimizer state, subsequent training again produces large changes in the output logit mean and further loss spike events.

Together with the context perturbations in \Cref{app:swaps}, this suggests that individual batches do not trigger loss spikes, but rather an underlying instability in the model's parameters and optimizer states. Consequently, we hypothesize that data filtering of this kind actually targets the symptom rather than the underlying source of the instability.

\FloatBarrier
\subsection{Learnable RMSNorm gains across seeds}
\begin{figure}[H]
  \centering
  \includegraphics[width=\columnwidth]{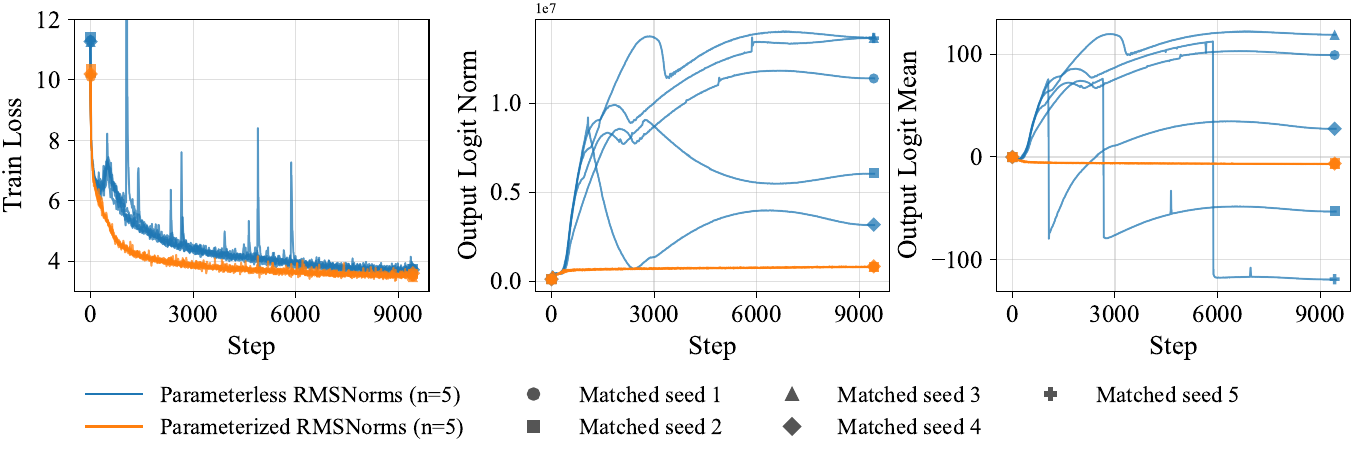}
  \caption{\textbf{Parameterized RMSNorms effectively mitigate fast loss spikes.} The five parameterized runs are highly consistent and overlap closely, indicating low seed-to-seed variability relative to the parameter-less runs.}
  \label{fig:rmsnorm_companion_pairs}
\end{figure}

\Cref{fig:rmsnorm_companion_pairs} extends the matched comparison in \Cref{fig:parameterized_vs_parameterless} to five seed pairs. Without learnable gains, every seed exhibits fast spikes accompanied by large swings in the output logit mean. With learnable gains, all five runs train smoothly, with much less growth/drift in the output logit norm and mean, and their trajectories nearly overlap.

\FloatBarrier
\subsection{Mitigations in the fast spike regime}
\label{sec:fast_spikes_mitigation_results}
We also evaluate the mitigations from \Cref{sec:mitigation_methods} in the fast spike regime, following the setup of \Cref{sec:characterization_fast}. For LM head SGD-M, we use momentum $\beta \in \{0.9, 0.99\}$ and apply no weight decay to the LM head. We additionally evaluate \emph{row-centering}, which subtracts each token's mean across the hidden dimension from its update ($M^t_{i,j} \leftarrow M^t_{i,j} - \frac{1}{d}\sum_{k=1}^d M^t_{i,k}$). Unlike $\mu$-centering, this operation is local to each vocabulary row and does not restore the zero-sum structure across the vocabulary.

\paragraph{Why SGD-M requires higher momentum.}
On the LM head, Adam's coordinate-wise rescaling amplifies updates to rows of rare tokens, whose second-moment estimates are small, relative to the updates SGD would produce \citep{stollenwerk-stollenwerk-2025-better}. This adaptivity has been identified as a key reason Adam outperforms gradient descent on language modeling \citep{kunstner2024heavy}. Replacing AdamW with SGD-M removes it, which substantially reduces the effective step size for most rows. Increasing the momentum coefficient from $0.9$ to $0.99$ raises the steady-state amplification factor $1/(1-\beta)$ from $10\times$ to $100\times$, and we find that the higher value is important in practice: $\beta = 0.99$ outperforms $\beta = 0.9$ at every learning rate and both model sizes in \Cref{tab:lr_sweep_spike_prone_regime}. This may explain why \citet{stollenwerk-stollenwerk-2025-better}, who evaluate SGD-M on the LM head with $\beta = 0.9$, find it slightly worse than their proposed coupled Adam.

\begin{figure}[H]
  \centering
  \includegraphics[width=.9\columnwidth]{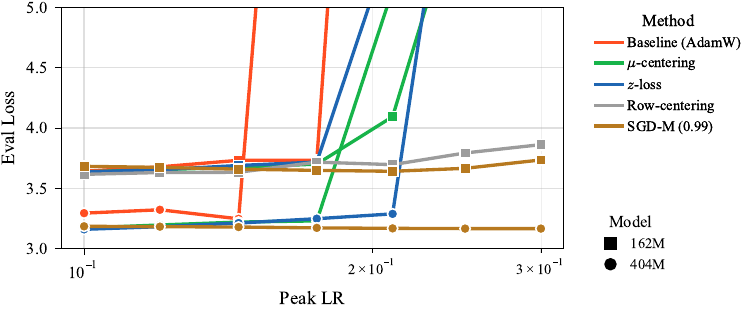}
  \caption{\textbf{LM head SGD-M is the only intervention stable across the spike-prone learning rate regime at both model sizes.} Final evaluation loss versus peak learning rate for 162M and 404M models. $\mu$-centering and $z$-loss diverge at some $\eta \in [10^{-1}, 3 \times 10^{-1}]$ at both sizes, and row-centering, while stable at 162M, diverges inconsistently at 404M and is omitted there. LM head SGD-M remains stable throughout and achieves the lowest loss at the high end.}
  \label{fig:lr_sensitivity}
\end{figure}

\paragraph{Learning rate sweep.}
We compare the mitigation methods across learning rates $\eta \in [10^{-1}, 3\times10^{-1}]$, spanning the regime in which fast spikes emerge. As shown in \Cref{fig:lr_sensitivity}, the methods differ substantially in both the range over which they suppress spikes and their final evaluation loss. $\mu$-centering and $z$-loss prevent fast spikes over much of the sweep but become unstable at higher learning rates, and row-centering remains stable at 162M but diverges inconsistently at 404M. LM head SGD-M with $\beta = 0.99$ remains stable across the full range at both model sizes and frequently attains the lowest final evaluation loss at higher learning rates. Full numerical results are reported in \Cref{tab:lr_sweep_spike_prone_regime}.

\begin{table}
  \centering
  \caption{\textbf{Learning rate sweep across the spike-prone regime ($\eta \in [10^{-1},\, 3{\times}10^{-1}]$) for 162M and 404M models.}
    Final evaluation loss ($\downarrow$ lower is better) is reported.
  $\dagger$ marks runs whose loss spiked during training and did not recover.}
  \label{tab:lr_sweep_spike_prone_regime}
  \small
  \begin{tabular}{lccccccc}
    \toprule
    & \multicolumn{7}{c}{\textbf{Peak learning rate $\eta$} \quad (final eval loss $\downarrow$)} \\
    \cmidrule(lr){2-8}
    \textbf{Method} & $10^{-1}$ & $1.2{\cdot}10^{-1}$ & $1.45{\cdot}10^{-1}$ & $1.75{\cdot}10^{-1}$ & $2.1{\cdot}10^{-1}$ & $2.5{\cdot}10^{-1}$ &
    $3{\cdot}10^{-1}$ \\
    \midrule
    \multicolumn{8}{l}{\emph{162M}} \\
    \midrule
    AdamW & 3.642 & 3.679 & 3.733 & 3.733 & $12.552^{\dagger}$ & $23.223^{\dagger}$ & $10.845^{\dagger}$ \\
    $\mu$-centering & 3.637 & 3.656 & 3.674 & 3.701 & $4.094^{\dagger}$ & $6.063^{\dagger}$ & $5.950^{\dagger}$ \\
    $z$-loss & 3.636 & 3.655 & 3.690 & 3.718 & $5.594^{\dagger}$ & $5.964^{\dagger}$ & $7.495^{\dagger}$ \\
    row-centering & \textbf{3.617} & \textbf{3.631} & \textbf{3.633} & 3.718 & 3.697 & 3.793 & 3.863 \\
    SGD-M, $\beta{=}0.9$ & 3.926 & 3.891 & 3.869 & 3.849 & 3.821 & 3.803 & 3.789 \\
    SGD-M, $\beta{=}0.99$ & 3.683 & 3.674 & 3.659 & \textbf{3.649} & \textbf{3.642} & \textbf{3.666} & \textbf{3.736} \\
    \midrule
    \multicolumn{8}{l}{\emph{404M}} \\
    \midrule
    AdamW & 3.295 & 3.323 & 3.248 & $11.125^{\dagger}$ & $11.006^{\dagger}$ & $16.731^{\dagger}$ & $28.241^{\dagger}$ \\
    $\mu$-centering & 3.172 & 3.197 & 3.221 & 3.231 & $5.144^{\dagger}$ & $6.088^{\dagger}$ & $7.115^{\dagger}$ \\
    $z$-loss & \textbf{3.161} & 3.183 & 3.213 & 3.249 & 3.289 & $7.266^{\dagger}$ & $6.588^{\dagger}$ \\
    row-centering & 3.257 & \textbf{3.166} & 3.550 & $4.930^{\dagger}$ & $12.106^{\dagger}$ & 3.310 & 3.452 \\
    SGD-M, $\beta{=}0.9$ & 3.335 & 3.319 & 3.304 & 3.280 & 3.268 & 3.255 & 3.244 \\
    SGD-M, $\beta{=}0.99$ & 3.186 & 3.183 & \textbf{3.180} & \textbf{3.173} & \textbf{3.169} & \textbf{3.167} & \textbf{3.167} \\
    \bottomrule
  \end{tabular}
\end{table}

\paragraph{Spike frequency across seeds.}
\label{para:spike_score}
To compare stability across runs, we follow
\citet{olmo20252olmo2furious} and define the \emph{spike score} as the
percentage of training steps whose loss exceeds seven standard deviations
from a trailing 1000-step window. \Cref{tab:main_results} reports both
spike score and final evaluation loss across seeds and model scales,
allowing us to distinguish suppression of visible spikes from differences
in final evaluation loss. The mitigation methods reduce fast spike
incidence to differing degrees, with LM head SGD-M exhibiting low spike
scores across the settings considered.

\begin{table}[h]
  \centering
  \small
  \caption{\textbf{Tuned SGD-M on the LM head suppresses spikes and remains competitive on evaluation loss.} Final evaluation loss and training-loss spike score (defined in \Cref{para:spike_score}) at peak learning rate $\eta = 0.15$ for 162M and 404M models. Values are mean $\pm$ standard error across 5 seeds; bold marks the lowest evaluation loss for each model scale.}
  \label{tab:main_results}
  \resizebox{\linewidth}{!}{
    \begin{tabular}{lcccc}
      \toprule
      & \multicolumn{2}{c}{\textbf{162M}} & \multicolumn{2}{c}{\textbf{404M}} \\
      \cmidrule(lr){2-3} \cmidrule(lr){4-5}
      \textbf{Method} & Eval Loss $\downarrow$ & Spike Score $\downarrow$ & Eval Loss $\downarrow$ & Spike Score $\downarrow$ \\
      \midrule
      Baseline (AdamW)                       & $3.702 \pm 0.014$ & $0.088 \pm 0.029$ & $3.916 \pm 0.278$ & $0.085 \pm 0.029$ \\
      $z$-loss                   & $3.644 \pm 0.006$ & $0.064 \pm 0.027$ & $3.214 \pm 0.008$ & $0.106 \pm 0.040$ \\
      $\mu$-centering            & $3.646 \pm 0.009$ & $0.066 \pm 0.043$ & $3.208 \pm 0.004$ & $0.064 \pm 0.002$ \\
      Row-centering    & $\mathbf{3.621 \pm 0.016}$ & $0.038 \pm 0.007$ & $4.858 \pm 1.438$ & $0.056 \pm 0.011$ \\
      SGD-M, $\beta=0.9$ & $3.855 \pm 0.003$ & $0.038 \pm 0.004$ & $3.301 \pm 0.003$ & $0.060 \pm 0.002$\\
      SGD-M, $\beta=0.99$ & $3.644 \pm 0.005$ & $0.038 \pm 0.004$ & $\mathbf{3.178 \pm 0.002}$ & $0.069 \pm 0.003$\\
      \bottomrule
    \end{tabular}
  }
\end{table}

\begin{figure}[htbp]
  \centering
  \includegraphics[width=\columnwidth]{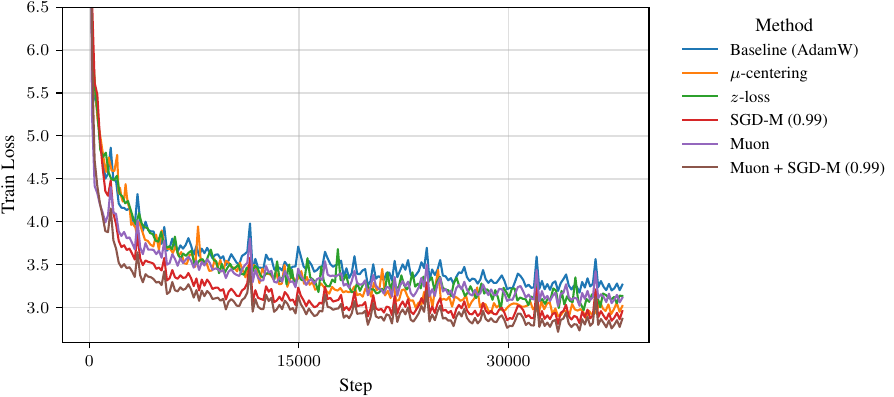}
  \caption{\textbf{LM head SGD-M also stabilizes training at 1.4B scale and pairs well with Muon.} Training loss curves for 1.4B models at peak learning rate $\eta = 0.07$ (single seed). Replacing AdamW with SGD-M ($\beta = 0.99$) on the LM head reaches a lower final training loss than AdamW, $\mu$-centering, and $z$-loss, and combining a Muon backbone with LM head SGD-M reaches the lowest loss of any configuration shown.}
  \label{fig:gpt3xl_runs}
\end{figure}

\paragraph{Scaling to 1.4B parameters.}
We additionally repeat the comparison at 1.4B scale. As shown in
\Cref{fig:gpt3xl_runs}, the methods again differ in their behavior at
larger scale; in this experiment, LM head SGD-M reaches a lower training
loss than AdamW, $\mu$-centering, and $z$-loss. We report the corresponding
results with a Muon~\citep{jordan2024muon} backbone in the same figure.

\FloatBarrier
\section{Additional time-bomb results}

\subsection{Full learning rate sweep}

\begin{figure}[H]
  \centering
  \includegraphics[width=\columnwidth]{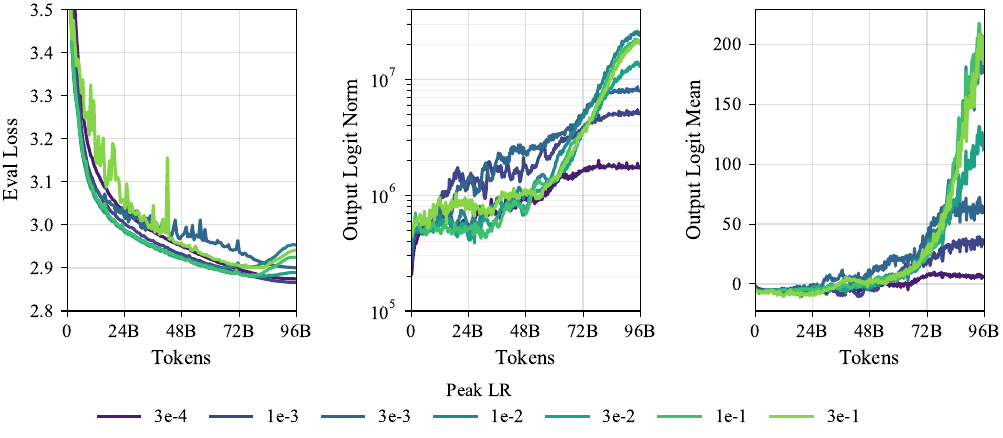}
  \caption{\textbf{Time-bomb instability emerges across the long-horizon learning rate sweep alongside accelerating output logit growth.}
  Evaluation loss (left), output logit norm (middle), and output logit mean (right) for 404M-parameter models trained for 96B tokens across peak learning rates from $3\times10^{-4}$ to $3\times10^{-1}$. Runs train similarly for much of pretraining, while higher-learning rate settings exhibit substantially greater late stage logit norm growth and, in several cases, deterioration in evaluation loss.}
  \label{fig:full_time_bomb}
\end{figure}

\Cref{fig:full_time_bomb} shows the complete 404M learning rate sweep previewed in \Cref{fig:time_bomb}. Runs at all learning rates follow similar loss trajectories for much of training, but higher learning rates produce substantially greater growth/drift in both the output logit norm and mean, accompanied in several cases by time-bomb spikes.

\FloatBarrier
\subsection{Effect of training horizon}

\begin{figure}[H]
  \centering
  \includegraphics[width=\linewidth]{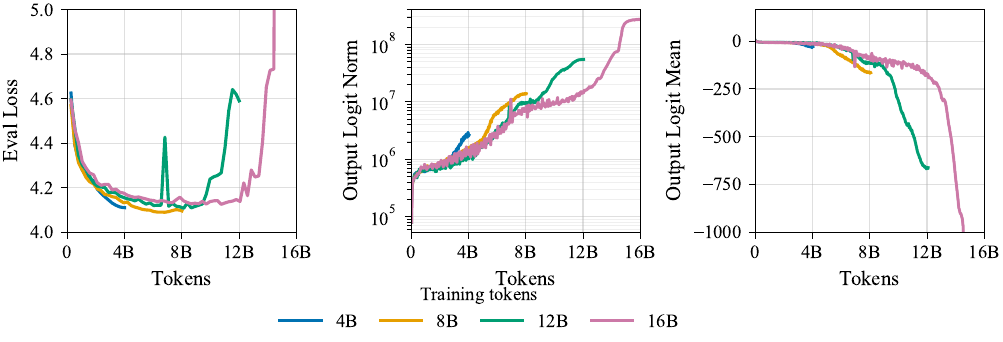}
  \caption{\textbf{Longer training horizons are associated with increasingly severe late stage LM head growth.}
  Evaluation loss (left), output logit norm (middle), and output logit mean (right) for 28M parameter models with parameterized RMSNorm, and peak LR of 0.1 trained with horizons of 4B, 8B, 12B, and 16B tokens. Extending training exposes progressively larger output logit norm growth and late stage loss deterioration, consistent with time-bomb instability becoming more prominent during prolonged training.}
  \label{fig:token_horizon}
\end{figure}

\Cref{fig:token_horizon} examines how the training horizon affects time-bomb instability in 28M-parameter models. Shorter runs end before severe late stage behavior appears, while longer runs show increasingly large output logit norms, substantial drift in the mean logit, and eventual deterioration in evaluation loss. Time-bomb instabilities can therefore remain hidden in shorter training runs and emerge only as training is extended.

\FloatBarrier
\subsection{LM head tuning details}\label{app:lm_head_hyperparam}

Because LM head SGD-M uses a separately tuned learning rate and momentum
coefficient, we additionally tune LM head specific AdamW hyperparameters. We consider tuning the LM head learning rate independently of the trunk learning rate (using the grid detailed in \cref{sec:time_bomb_overtraining}), increasing the LM head first moment coefficient from the default $\beta_1=0.9$ to $\beta_1=0.99$, and combining these changes with $\mu$-centering and $z$-loss. As shown in
\Cref{fig:mitigations_tuned_400M}, tuning the LM head learning rate alone
provides only modest improvement, whereas increasing $\beta_1$ produces a
substantially larger gain in final evaluation loss.

These runs were performed across three hardware configurations:
$4\times$ NVIDIA B200, $8\times$ NVIDIA H100, and $8\times$ NVIDIA A100
GPUs.

\FloatBarrier
\section{Precision analysis}

\paragraph{Precision settings.}
\label{app:precision_settings}
In BF16 mixed precision, parameters and optimizer states are stored in FP32, while weights and activations are cast to BF16 at runtime so that matrix multiplications run on BF16 hardware units. In Flax, this corresponds to \texttt{param\_dtype=float32} with \texttt{dtype=bfloat16}, analogous to \texttt{torch.autocast} in PyTorch. BF16 true stores the LM head weights and optimizer states in BF16, removing the FP32 master copies. In the FP32 setting, both the LM head weights and the LM head GEMM are kept in full FP32 precision. All three settings use the default GEMM routine for their respective precision, and all non-LM-head layers use BF16 mixed precision throughout.

\subsection{Why error grows with the logit mean}
\label{app:precision_analysis}

\begin{figure}[t]
  \centering
  \includegraphics[width=0.8\columnwidth]{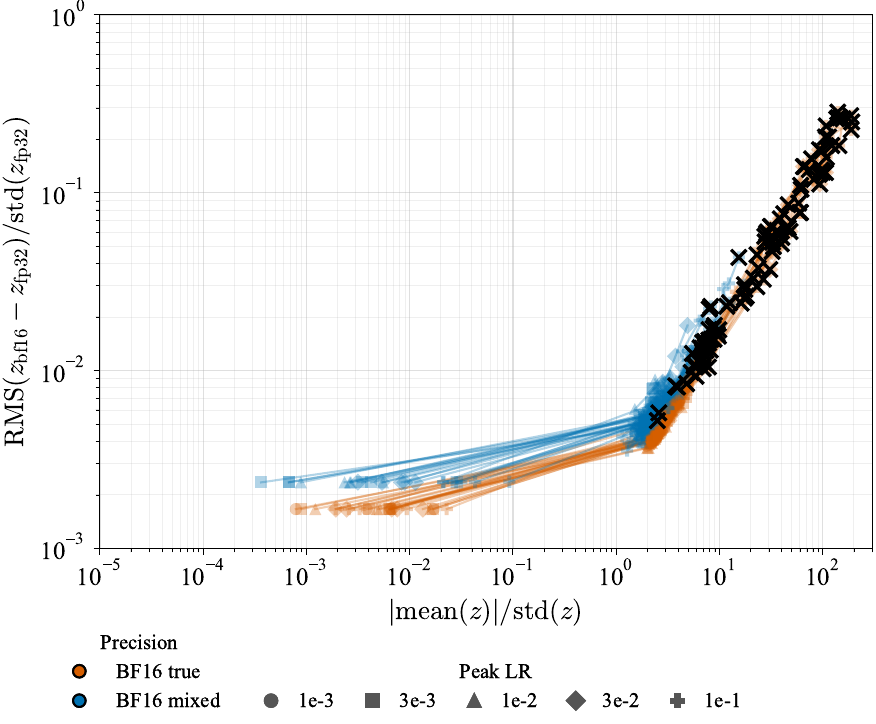}
  \caption{\textbf{Normalized logit error grows as the logit mean becomes
    large relative to variation across logits.}
    Across BF16 true and BF16 mixed baseline runs, the normalized RMS error
    between native BF16 logits and FP32 reference logits,
    $\mathrm{RMS}(z_{\mathrm{bf16}}-z_{\mathrm{fp32}})
    / \mathrm{std}(z_{\mathrm{fp32}})$,
    increases with the relative logit mean,
    $|\mathrm{mean}(z)|/\mathrm{std}(z)$.
    Colors indicate precision, markers indicate peak learning rate, and lines
  trace individual training trajectories.}
  \label{fig:precision_v_logit}
\end{figure}

In \Cref{sec:precision_accelerates_time_bomb}, we showed that quantization error in the LM head grows alongside the output logits before time-bomb spikes. Here we sketch why this happens, based on the fixed relative precision of BF16.

BF16 has seven stored fraction bits and an implicit leading bit, giving eight bits of significand precision \citep{kalamkar2019studybfloat16deeplearning}. For normal values $x \in [2^e, 2^{e+1})$, the spacing between consecutive representable values is
\[
  \operatorname{ULP}(x) = 2^{e-7}.
\]
This spacing grows with magnitude; between $1$ and $2$, adjacent BF16 values are approximately $0.008$ apart, whereas between $1024$ and $2048$, they are $8$ apart.

A large logit mean can increase the magnitude of the logits without changing the differences that determine softmax probabilities. Because BF16 spacing grows with magnitude, these differences can become too small to represent accurately.

As a concrete example, the logits $(0, 1)$ yield probabilities of approximately $(0.27, 0.73)$. Shifting both by $-1025$ leaves the exact probabilities unchanged, but because adjacent BF16 values near 1024 are 8 apart, both $-1025$ and $-1024$ round to $-1024$, collapsing the distribution to $(0.5, 0.5)$. Numerically stable softmax implementations, which subtract the maximum logit, cannot recover this difference once it has been lost in the computation that produced the logits.

Because rounding error scales with the magnitude of each logit, while the softmax depends only on the differences between logits, the relevant quantity is the logit mean relative to the spread of the logits,
\[
  r = \frac{|\mu_z|}{\sigma_z}.
\]

When $r \ll 1$, typical logit magnitudes are
comparable to $\sigma_z$, so the BF16 rounding
resolution is small relative to the spread of the
logits. When $r \gg 1$, the logit magnitudes are
instead dominated by $|\mu_z|$, and numerical errors
can become much larger relative to the differences
that determine the softmax distribution.

As shown in \Cref{fig:precision_v_logit}, the
normalized numerical error
\[
  \frac{\mathrm{RMS}(z_{\mathrm{bf16}} - z_{\mathrm{fp32}})}
  {\mathrm{std}(z_{\mathrm{fp32}})}
\]
increases as $r$ grows across both BF16 precision
settings and the learning rates tested. This trend
is consistent with the loss of numerical resolution
predicted by BF16's fixed relative precision.

\FloatBarrier
\subsection{Lower precision triggers time bombs at smaller learning rates}

\begin{figure}[H]
  \centering
  \includegraphics[width=0.8\columnwidth]{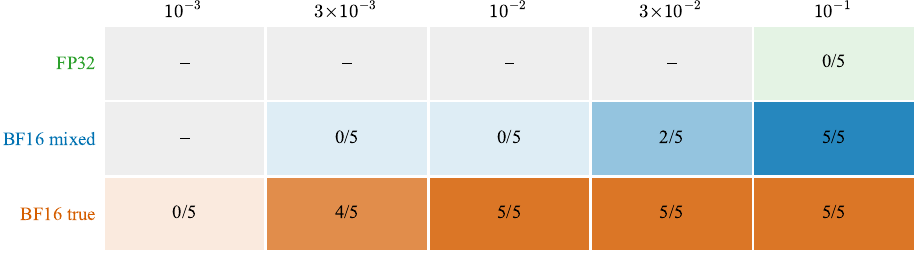}
  \caption{\textbf{Lower precision triggers time-bomb spikes at smaller learning rates.}
  Fraction of seeds exhibiting a time-bomb instability across peak learning rates and numerical precision settings. As precision decreases from FP32 to BF16 mixed to BF16 true, instability emerges at progressively smaller learning rates. A time-bomb is defined as three consecutive post-midpoint evaluations at least $0.01$ above the running-best evaluation loss.}
  \label{fig:precision_time_bomb_boundary}
\end{figure}

\Cref{fig:precision_time_bomb_boundary} shows the fraction of seeds that develop time-bomb spikes across the full learning rate sweep for the 28M precision experiments in \Cref{sec:precision_accelerates_time_bomb}. Reducing precision causes time-bomb spikes to appear at smaller learning rates, and FP32 remains stable even at the largest learning rate, where every BF16 run fails.

\FloatBarrier
\subsection{Mitigations under BF16 true}

\begin{figure}[H]
  \centering
  \includegraphics[width=\columnwidth]{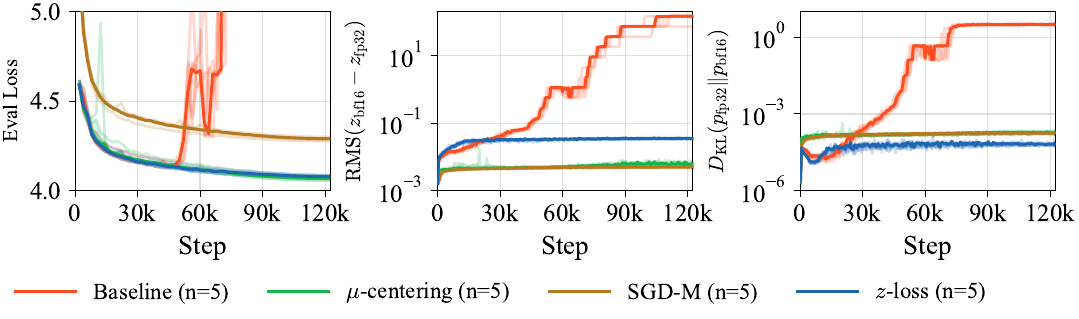}
  \caption{\textbf{Methods that prevent time-bomb spikes also suppress quantization error.} For 28M-parameter models trained in BF16 true at peak learning rate \(0.1\), the unmitigated baseline develops severe unrecoverable loss spikes (left) together with orders-of-magnitude growth in logit quantization error (middle) and predictive distribution quantization error (right). $\mu$-centering, $z$-loss, and LM head SGD-M prevent the time bomb and keep both numerical-error metrics bounded across five seeds. Solid lines show medians and faint lines show individual runs.}
  \label{fig:mitigations_30M}
\end{figure}

\Cref{fig:mitigations_30M} repeats the mitigation comparison of \Cref{fig:mitigations_28M_bf16_mixed} under BF16 true, where the LM head weights and optimizer states are also stored in BF16. The same pattern holds and is more pronounced: the unmitigated baseline fails after quantization error grows by several orders of magnitude, while $\mu$-centering, $z$-loss, and LM head SGD-M all keep the error bounded and prevent the time bomb. In this setting, however, LM head SGD-M converges to a noticeably higher loss than the other mitigations, even though its quantization error remains low. The gap appears early in training rather than as a late failure, suggesting slower optimization rather than instability.

\FloatBarrier
\subsection{Comparison with the fixed softmax of \citet{barron2026fixsoftmaxcrossentropy}}
\label{app:softmax_fix_results}

Concurrent work shows that standard softmax cross entropy implementations can round the target logit's gradient to zero at high confidence, breaking the zero-sum structure of the cross entropy gradient with respect to the logits \citep{barron2026fixsoftmaxcrossentropy, liu2026grokkingglitchinglowprecisiondrives}. To test whether this Softmax Collapse contributes to time-bomb spikes, we replace standard cross entropy with the fixed implementation of \citet{barron2026fixsoftmaxcrossentropy}, which preserves the zero-sum structure at large logit margins. Importantly, this correction addresses numerical errors in the cross entropy computation and its gradients, but does not directly eliminate the mean drift induced by Adam's preconditioner or recover logit differences lost during the BF16 forward computation of the LM head.

\begin{figure}[H]
  \centering
  \includegraphics[width=\columnwidth]{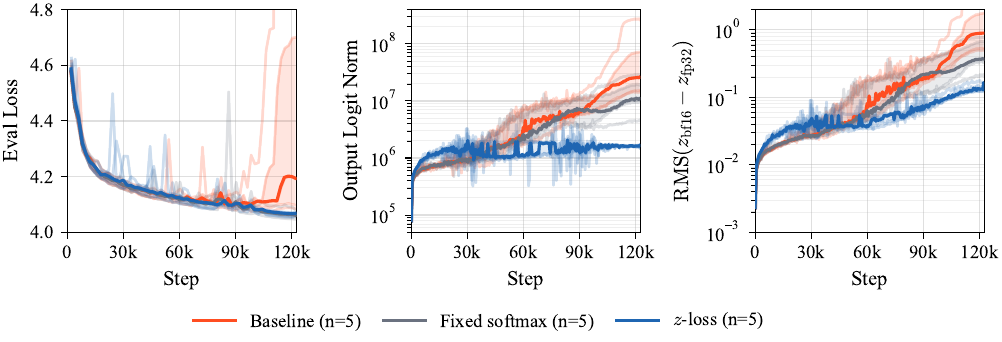}
 \caption{\textbf{Under BF16 mixed, the fixed softmax avoids divergence within the training horizon, but logit growth and quantization error persist.} For 28M-parameter models trained at peak learning rate $0.1$, we compare standard cross entropy, the fixed softmax of \citet{barron2026fixsoftmaxcrossentropy}, and $z$-loss. The fixed softmax tracks the baseline for most of training, and its output logit norm (middle) and quantization error (right) remain significantly higher than $z$-loss runs. Solid lines show medians across five seeds and faint lines show individual runs.}

  \label{fig:softmax_fix_bf16_mixed}
\end{figure}

\begin{figure}[H]
  \centering
  \includegraphics[width=\columnwidth]{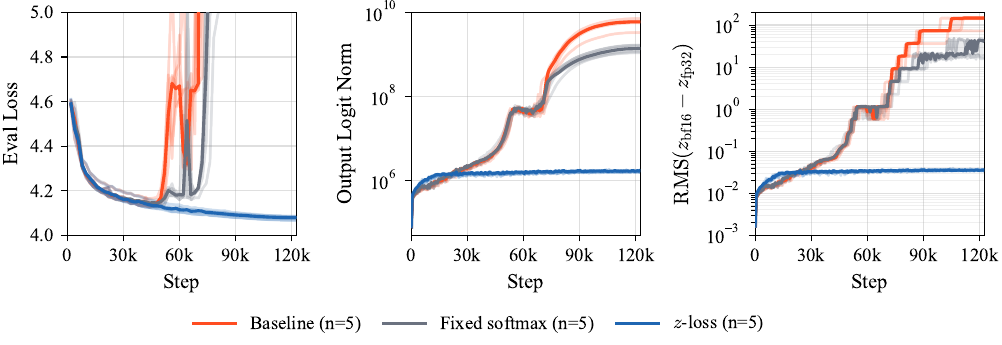}
  \caption{\textbf{Under BF16 true, the fixed softmax delays but does not prevent time-bomb spikes.} Same setup as \Cref{fig:softmax_fix_bf16_mixed}, with the LM head weights and optimizer states stored in BF16. The output logit norm and quantization error with the fixed softmax closely track the baseline until late in training, with five seeds still diverging, while $z$-loss keeps both bounded. Solid lines show medians across five seeds and faint lines show individual runs.}
  \label{fig:softmax_fix_bf16_true}
\end{figure}

We evaluate the corrected cross entropy against standard
cross entropy and $z$-loss in the 28M-parameter time-bomb
setting, using a peak learning rate of 0.1 and five seeds
per method. We consider both BF16 mixed and BF16 true
precision, as defined in \Cref{app:precision_settings}. Results are shown in \Cref{fig:softmax_fix_bf16_mixed,fig:softmax_fix_bf16_true}. Under BF16 true, the fixed softmax delays failure, but all five seeds still develop time-bomb spikes. Under BF16 mixed, none of the fixed softmax runs develops result in a time bomb spike over the 16B token training horizon. Nevertheless, output logit norms and quantization error continue to grow and remain substantially higher than runs with $z$-loss. 

Together, these results show that correcting cross
entropy can substantially improve training stability
without eliminating LM head logit growth or its
associated quantization error. However, the correction
alone is insufficient to prevent time-bomb instability
under BF16 true precision.

\newpage
\end{document}